\documentclass[letterpaper, 10 pt, conference]{ieeeconf}  

\IEEEoverridecommandlockouts                              

\usepackage{cite}
\usepackage{svg}

\usepackage{amsmath,amssymb,amsfonts}
\usepackage{algorithmic}
\usepackage{graphicx}
\usepackage{subcaption}
\usepackage{textcomp}
\usepackage{xcolor}
\usepackage{makecell}
\usepackage[font=small,labelfont=bf]{caption}
\def\BibTeX{{\rm B\kern-.05em{\sc i\kern-.025em b}\kern-.08em
		T\kern-.1667em\lower.7ex\hbox{E}\kern-.125emX}}

\title{\Large \bf 
	Domain-Independent Disperse and Pick method for Robotic Grasping}

\author{Prem Raj$^*$$^1$, Aniruddha Singhal$^*$$^2$, Vipul Sanap$^2$ , L. Behera$^{1,2}$ and Rajesh Sinha$^2$
	\thanks{$^*$Equal contribution}
	\thanks{$^1$
		Intelligence Systems and Control (ISCon) Lab \textit{Indian Institute of Technology} Kanpur, India praj@iitk.ac.in, lbehera@iitk.ac.in}
	\thanks{$^2$ Smart Machines \textit{TCS Research} Noida, India. aniruddha.singhal@tcs.com, vipul.sanap1@tcs.com, rajesh.sinha@tcs.com}}

\begin{document}
	
	\maketitle
	
	\begin{abstract}
	
	    Picking unseen objects from clutter is a difficult problem because of the variability in objects (shape, size, and material) and occlusion due to clutter. As a result, it becomes difficult for grasping methods to segment the objects properly and they fail to singulate the object to be picked. This may result in grasp failure or picking of multiple objects together in a single attempt. A push-to-move action by the robot will be beneficial to disperse the objects in the workspace and thus assist the grasping and vision algorithm. We propose a disperse and pick method for domain-independent robotic grasping in a highly cluttered heap of objects. The novel contribution of our framework is the introduction of a heuristic clutter removal method that does not require deep learning and can work on unseen objects. 
        At each iteration of the algorithm, the robot either performs a push-to-move action or a grasp action based on the estimated clutter profile.
		For grasp planning, we present an improved and adaptive version of a recent domain-independent grasping method.
		The efficacy of the integrated system is demonstrated in simulation as well as in the real-world.

	\end{abstract}
	
	%
	%

	\section{Introduction}
	 A truly universal robotic grasping solution should be able to grasp a wide variety of novel objects in a time-effective manner with minimal reconfiguration. The use of robotics is now not just limited to industrial or highly structured settings but has extended to domestic and unstructured environments. The next phase of automation requires robots to work in more challenging and dynamic environments \cite{stoica2017berkeley}.
	For example, the e-commerce industry has seen a massive adoption of robotics due to the increasing load. 

	There has been a huge interest in automating the process of grasping and plenty of interesting solutions were seen in the series of robotics picking challenges organized by e-commerce giant Amazon \cite{amazon_picking_challenge}. Most of the solutions present in APC were learning based as it was essential to recognize the item before picking. However, those solutions were not extendable to a large number of unseen objects and also often faced difficulty in picking items from a clutter. This particular problem of picking unseen items from clutter has gained sufficient traction from the industry due to its wide applicability across domains.
	Also, because of short product life cycles and mass-personalization, a robot must be able to pick a new item in any environment with minimal reconfiguration \cite{reinhart2011automatic} \cite{el2019simulation}. This problem of robotic grasping has huge applications in industrial, domestic, medical, and retail settings. 
		For example, this problem can be seen in manufacturing plants where several parts come in a tray and need to be put in specific places. It is also seen in retail stores where a clutter of objects needs to be cleared before restocking items on a shelf. This is one of the key problems which does not allow full-scale automation in retail stores because a human is needed to clear the clutter of objects. If the clutter is removed automatically then the rest of the process of stocking the items on the shelves can be fully automated.
		\begin{figure}
    	    \centering
        	\begin{subfigure}{0.15\textwidth}
        		\centering
        		\includegraphics[width=\textwidth]{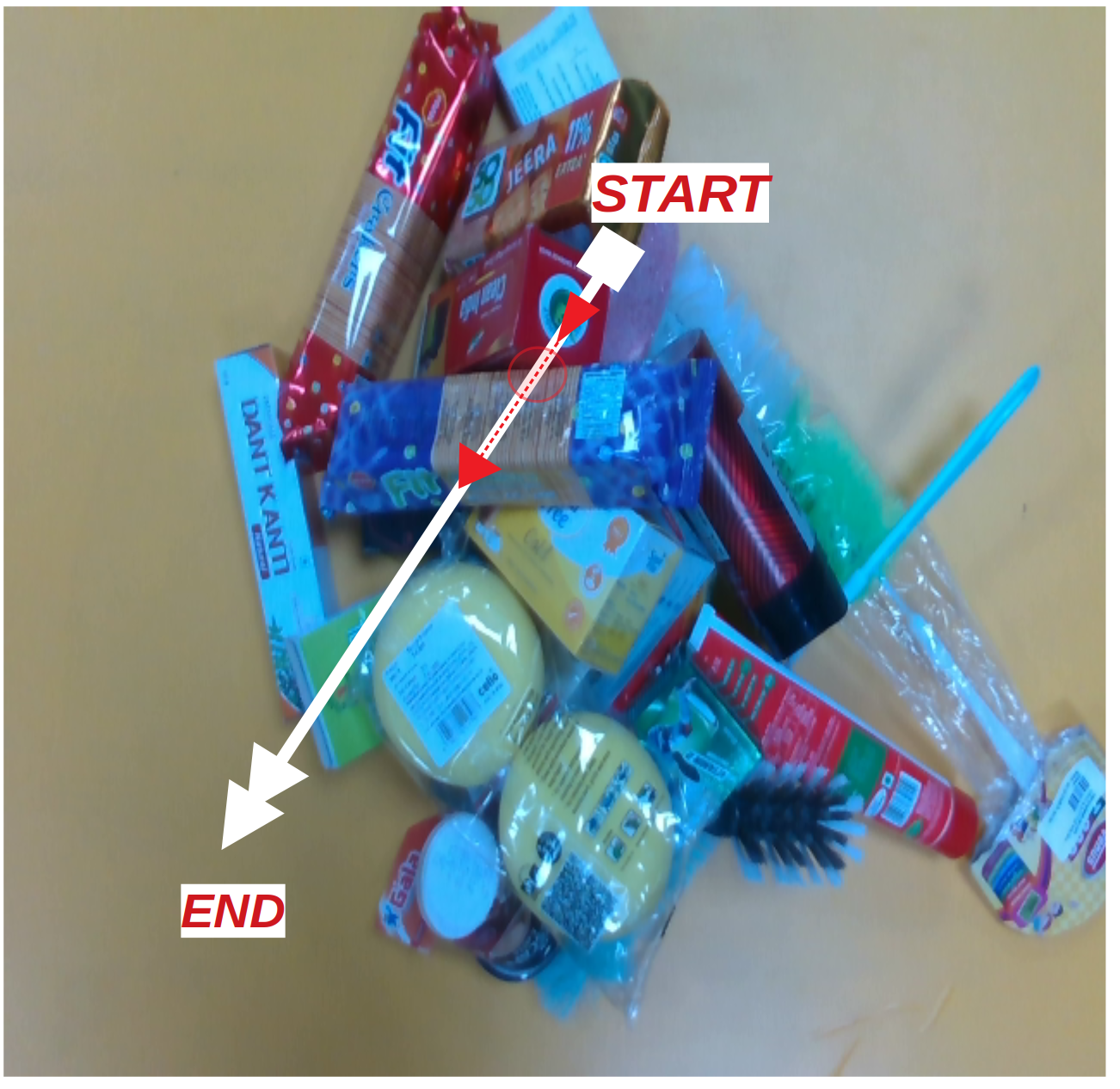}
        		\caption{Predicted push action}
        		\label{fig:decluttering}
        	    \end{subfigure}
        	 \begin{subfigure}{0.15\textwidth}
        		\centering
        		\includegraphics[width=0.82\textwidth]{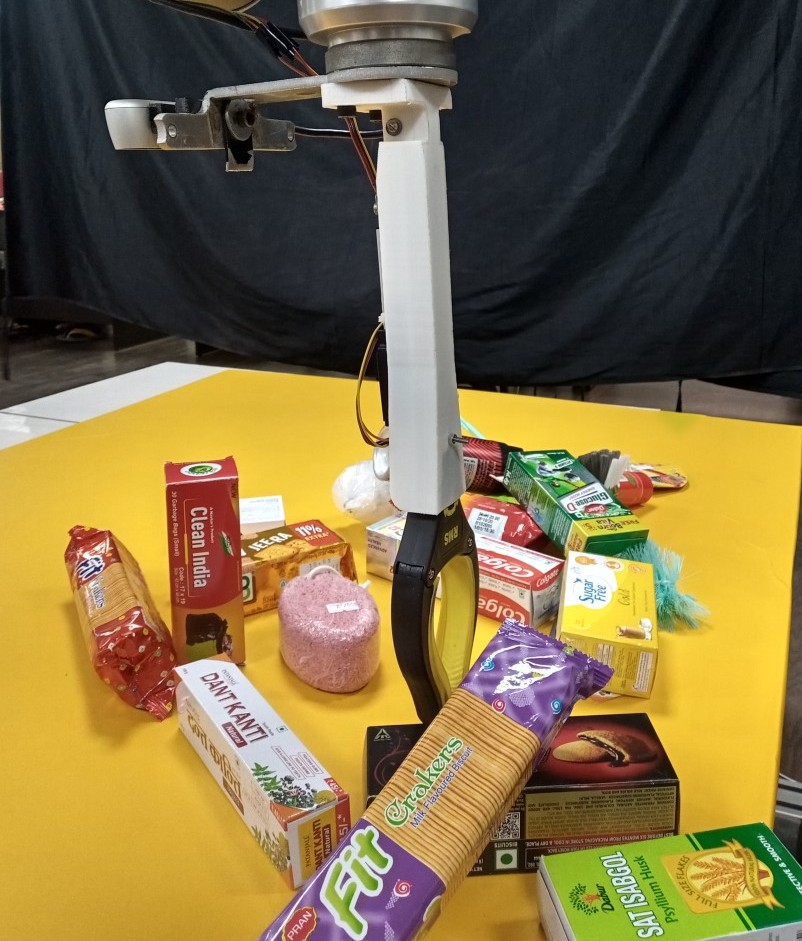}
        		\caption{Robot pushing the clutter}
        		\label{fig:free_point_space}
    	    \end{subfigure}
    		\begin{subfigure}{0.15\textwidth}
        		\centering
        		\includegraphics[width=\textwidth]{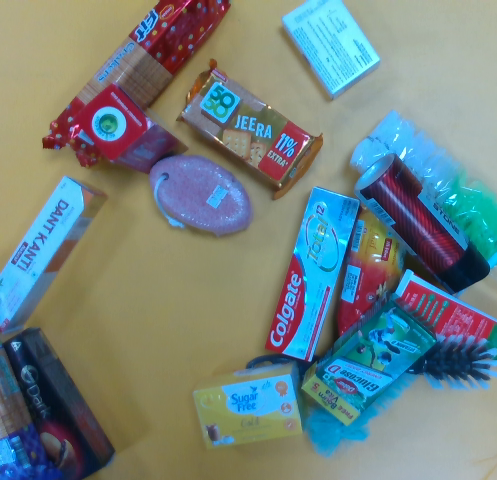}
        		\caption{Decluttered objects post action}
    		    \label{fig:free_point_space}
        	\end{subfigure}

	    \caption{Effect of the push-to-move action. The objects after this action can be better segmented and singulated thereby assisting the grasp. It significantly reduces the cases of multi-picks (picking more than one object in a single grasp attempt). 
	   }
	    \label{fig:my_label}
	\end{figure}
	
	\begin{figure*}[h!]
		\centering
		\includegraphics[width = \linewidth]{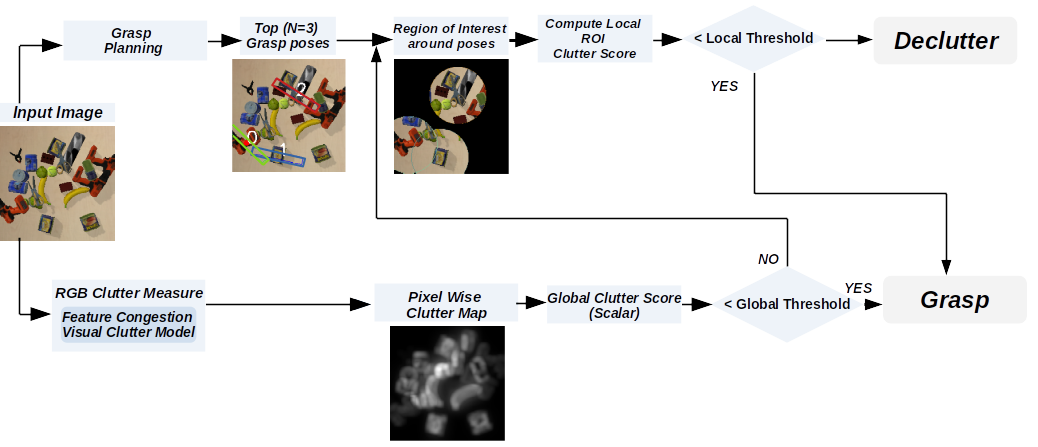}
		\caption{Depiction of the core implementation pipeline of our proposed method.}
		\label{fig:overview}
	\end{figure*}
	


	In this paper, we propose a solution to the problem of robotic grasping wherein we develop a framework to declutter the environment (disperse the objects) when required based on the amount of clutter and pick the items from the tabletop stably. The proposed method is a model-free (no CAD model of the objects) heuristic method which can be used for unfamiliar, rigid objects of daily use. The main contributions of the paper are summarised as follow:
	\begin{itemize}
	    \item A domain-independent lightweight framework for clutter quantification and estimation and a strategy to manipulate the environment to disperse clutter. As a result, the objects are properly segmented by the vision algorithm and are thus grasped with increased stability and certainty.
	    \item An improved and adaptive version of a recent domain-independent grasp planning algorithm \cite{pharswan2019domain} 
	    \item A simulation setup is created that implements the proposed method. It will be made available publicly to the research community for testing and further research.
	\end{itemize}

	

	\section{Literature Review}
	The standard methods used to solve the robotic grasping problem involve a vision module that can segment the object to be picked and a grasp pose generation algorithm \cite{kleeberger2020survey}. The vision module can either be analytic (i.e. requires no training on data) \cite{pharswan2019domain,jo2020object,vohra2019real} or empirical (i.e. requires training on data) \cite{dexnet2, dexnet4}. 
	Recently, deep learning has become a popular learning-based method to solve robotic grasping problem \cite{tremblay2018deep,levine2018learning,kalashnikov2018qt, james2019sim,redmon2015real,kumar2019semi,zeng2020tossingbot,song2020grasping}. In the sense of universal robotic grasping, these approaches have one or more limitations. In \cite{levine2018learning, kalashnikov2018qt}, massive training data was collected with a large number of robot hours to train the deep network, which is practically not feasible all the time. To overcome this limitation, \cite{james2019sim} trained their deep network entirely with the simulation data via sim-to-sim domain adaptation technique. However, the model's generalization capability to real-world setup has only been shown for the real-world objects (and their arrangements) that closely resemble the simulation environment. Semi-supervised techniques \cite{kumar2019semi} and low-cost demonstrations \cite{song2020grasping} were also proposed to reduce the amount of training data, however, they still require a significant amount of effort in the overall training pipeline while the domain shift problem still persists. In \cite{tremblay2018deep}, 3D object models of the target objects are required priory to generate the synthetic training dataset. Some other approaches are designed for specific use-cases such as picking objects in isolation \cite{redmon2015real} and learning to throw arbitrary objects \cite{zeng2020tossingbot}.
	One other deep learning-based robotic grasping framework namely DexNet \cite{dexnet2, dexnet4} had specifically focused upon the universal robotic grasping problem and had shown state-of-the-art performance and generalization capabilities to varieties of daily used objects. They collected simulation training data using thousands of 3D meshes of the objects and trained the deep network entirely on the simulated depth images. 
	Recently, an analytical method \cite{pharswan2019domain} had been proposed that targets universal robotic grasping problems without using any sophisticated deep learning-based approach. The method does not require any 3D model of the objects and does not assume anything about the scene objects. In this, performance compared to the state-of-the-art had been shown for universal robotic grasping in a table-top setting. This method is used as our underlying grasp generation algorithm with certain key improvements, discussed further in the Sec. \ref{sec:method}.
	
	While these approaches had been successful in picking objects in clutter to a certain extent, the problem of collision of the gripper with nearby objects still remains a challenge \cite{pharswan2019domain,dexnet4}. It is essential to have enough space for the gripper jaws or fingers to enter without colliding or damaging nearby objects. Otherwise, the grasp attempt might fail, more than one object might be picked at a time, or it might damage the nearby objects. One particular method enables the picking of cluttered objects by reorienting the gripper while approaching an object  \cite{ kumar2019semi}. This approach is limited to deciding grasp action for objects near bin walls or corners and does not generalize for the objects in clutter placed at the center. This opens up the need to declutter the surface by nudging items a little bit just like a human would do. In \cite{zeng2018learning}, an attempt is made to learn the synergies between pushing and grasping using deep reinforcement learning, and experiments are done to pick regular geometric shaped objects arranged in a tight configuration and a limited range of other shapes (fruit, bottles, etc.), however, the generalization capabilities have not been shown to the objects of daily-use having varieties of (irregular) shapes and sizes. Similar work has proposed an interactive segmentation method by disturbing the scene a little bit with the help of the robot end-effector, however, it requires complex motion planning and significant compute \cite{patten_action_2018}, \cite{katz_clearing_2013}.
	 In static-push (constant contact with objects throughout pushing action) approaches of decluttering action primitives, works in literature differentiate on the selection of start and end points of the action primitive.
	 In \cite{danielczuk_linear_2018}, the starting point is selected by representing the segmented point cloud of the objects, approximating objects identity as a point at the center of mass of segmented clusters. In \cite{cluster-push}, objects are first grouped into clusters based on the geometric proximity and then a push action is planned analytically. This method assumes that the pushing surface is planar with a homogeneous friction coefficient and the objects are identifiable in the scene which is not the case given a clutter of novel objects with various shapes and sizes. Similarly, a dynamic-push (impulsed based controlled scattering of objects by collision) \cite{imran_singulation_2019} and a hybrid (static-dynamic) approach \cite{khan_scattering_2020} are also proposed that use an analytical method to estimate push velocity, starting point, and principal component analysis to define push vector on the segmented images. This method heavily depends on learning certain properties of the scene (i.e. coefficient of restitution, friction, etc.).
	 In \cite{gupta_using_2012}, appropriate action primitive (either pick, push or nudge) is taken according to the clutter state (i.e. uncluttered, cluttered, and piled-cluttered) of the particular image region. However, the paper did not discuss the methods to identify the clutter states. Additionally, the study is limited to Lego-like bricks.
	 
	 The clutter removal framework could benefit greatly if a method can be included in the framework that can estimate the clutter quantitatively for any arbitrary scene. In a different class of works \cite{berg_crowding_2009,yu_modeling_2014,rosenholtz_feature_2006}, attempts have been made on quantitatively estimating the clutter in close correlation to human perceptual behavior. In \cite{berg_crowding_2009}, authors define clutter as being proportional to the notion of crowding and returns a single clutter value for the whole image. In \cite{rosenholtz_feature_2006}, a method named feature-congestion-model (FCM) is proposed to quantify the clutter that uses the CIE $Lab$ color space representation of the scene image, combining the different features in a way that closely resemble the human model of clutter. 
	 We select feature-congestion-model \cite{rosenholtz_feature_2006} as our underlying module for clutter estimation. Unlike others \cite{berg_crowding_2009,yu_modeling_2014}, FCM provides a pixel-wise map of the clutter which is desirable for robotic applications as with the use of the pixel-wise clutter-map, clutter state of any sub-region of the image can be assessed and handled accordingly. 

	\section{Method Description}\label{sec:method}
	
	\subsection{Clutter detection and removal} \label{sec-clutter-description}
    The majority of the failure cases in the previous domain-independent unsupervised technique 	 \cite{pharswan2019domain} are due to the presence of clutter in the near vicinity of the object to be picked.
	To deal with the problem of clutter, a clutter detection and removal framework is included in the robotic grasping pipeline. At each run of the algorithm, the robot should decide either to perform a push-to-move action or a grasp action. For simplicity of understanding, the whole approach is divided into three components. First, a pixel-wise clutter-map is calculated using the FCM model. Then, based on the estimated clutter profile a feasible action is assigned to the robot. Finally, if the robot is assigned for a push-to-move action, a linear push policy is decided by the algorithm. Each of the three components is described below.

	\begin{figure*}[!t]
	\centering
	\begin{subfigure}{0.18\textwidth}
		\centering
		\includegraphics[width=\textwidth]{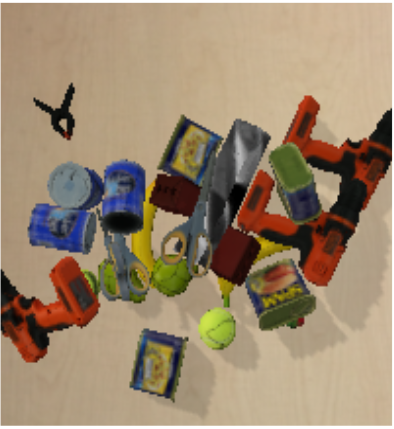}
		\caption{}
		\label{fig:decluttering}
	\end{subfigure}
	\hfill
	\begin{subfigure}{0.18\textwidth}
		\centering
		\includegraphics[width=\textwidth]{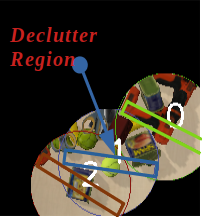}
		\caption{}
		\label{fig:ci_clutter_new}
	\end{subfigure}
	\hfill
	\begin{subfigure}{0.18\textwidth}
		\centering
		\includegraphics[width=\textwidth]{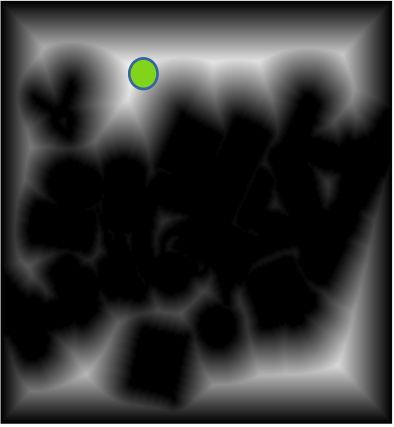}
		\caption{}
		\label{fig:ci_final_declutter}
	\end{subfigure}
	\hfill
	\begin{subfigure}{0.18\textwidth}
		\centering
		\includegraphics[width=1\textwidth]{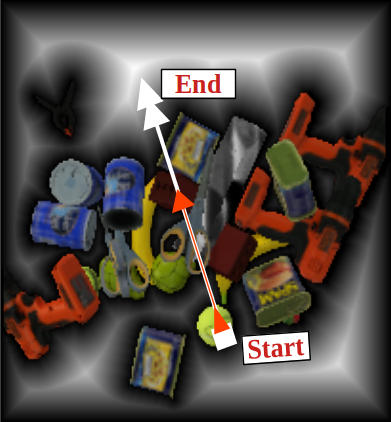}
		\caption{}
		\label{fig:ci_binary}
	\end{subfigure}
	\hfill
	\begin{subfigure}{0.178\textwidth}
		\centering
		\includegraphics[width=\textwidth]{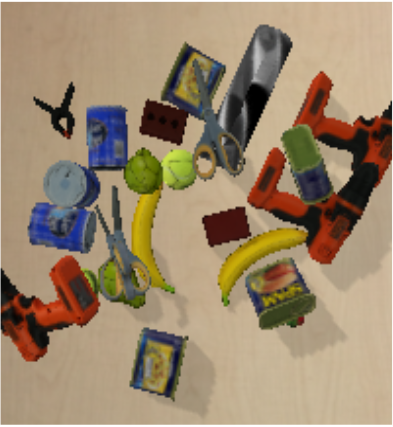}
		\caption{}
		\label{fig:free_point_space}
	\end{subfigure}
	\hfill
	
	\caption{Clutter Removal pipeline:(a) Initial state (b) ROI corresponds to the grasp pose 1 is selected for push-to-move action (c) Distance transform map and the freest point on the table top (shown in green). (d) Predicted push vector (e) Objects state post push-to-move action}
	\label{fig:ci_finalpl_declutter}
	
\end{figure*}

	\subsubsection{FCM model for pixel-wise clutter-map estimation}
	 Our chosen clutter model FCM \cite{rosenholtz_feature_2006} provides a pixel wise clutter map upon which a suitable strategy can be built to identify the cluttered regions around the grasping locations.
	The clutter metric estimates the degree of clutter for the target region of interest by calculating the local variability in certain key features such as color, contrast, and orientation. These features are known to be crucial constituents for the human perception model of clutter quantification \cite{rosenholtz_feature_2006}.
	To calculate the clutter metric, first, the image is transformed into CIE Lab color space with luminance channel $L$ and two chrominance channels $a$ and $b$. The CIE $Lab$ color space projects the image space akin to one occurring in human perception and poses high sensitivity in detecting small changes in color.
	 Luminance contrast is one of the features that is obtained by filtering the luminance channel $L$ with a difference-of-gaussians (DoG) filter. It serves as pattern segmentation and a measure of shape and size.
	 The orientation feature is also obtained from the luminance channel.
	 To obtain the orientation feature, the luminance channel is processed with four oriented edge detection filters of 0, 90, 45, and 135 degrees representing horizontal(H), vertical(V), and two (Left (L) and Right (R)) diagonals. The orientation feature plays a significant role by measuring the deviations in the texture.
	 To get the feature vectors for color space, all three channels namely $L$, $a$, and $b$ are used as it is. Though color is an important factor for defining the clutter, however, it may cause increased sensitivity of the model to the color. To make a trade-off, color features are down-weighted with a certain factor. For each feature, local variability is computed by performing point-wise (co)variance operation on the individual feature vectors. 
	 Finally, all the different feature maps are combined in a particular way, resulting in a pixel-wise clutter-map. Readers are advised to refer \cite{rosenholtz_feature_2006} for more details on different stages and processing involved in feature clutter map computation. An example of such a pixel-wise clutter map can be viewed in Fig. \ref{fig:overview}.

	\subsubsection{Deciding the feasible action}
    At each iteration of the algorithm, it decides between a grasp action and a push-to-move action. The decision algorithm is depicted in Fig. \ref{fig:overview}. Two types of reasoning are used as driving metrics for the decision algorithm. One is the local clutter score that represents the severity of clutter for the grasp pose region-of-interest (ROI) and another is the global clutter score for the whole image. Global clutter score is obtained by taking the average of pixel-wise clutter values over the whole image region. Local clutter is associated with each predicted grasp pose (we choose top $N=3$ grasp poses). It is obtained by taking the average of pixel-wise clutter values over a circular region around the grasp pose as shown in Fig. \ref{fig:overview}. The center of the ROI is the grasp pose center and the diameter is equivalent to the gripper opening length in the image plane. 
	If the global clutter score is less than a predefined threshold then the robot need not perform a push-to-move action and takes a grasp action according to the grasping algorithm. Otherwise, the top $N$ grasp poses are chosen and their respective local clutter scores are calculated. If each of the $N$ grasp locations has a clutter score greater than a threshold, then the robot performs a push-to-move action. Else, if any of the $N$ poses has a local clutter score less than the local threshold, a grasp action is decided. The one with the lowest local clutter score is chosen as the final grasp pose out of the top $N$ grasp poses. 
	To avoid repetitive failures, the algorithm decides to take a push-to-move action irrespective of the clutter state if three consecutive failures are encountered.   

\subsubsection{Clutter removal: push-to-move action}
    Once a region in the image is identified for the decluttering action by the algorithm, a suitable push vector needs to be calculated.
	 The push action is guided by a vector that has its endpoint chosen as the freest point in the robot workspace and its start point chosen in the vicinity of the selected cluttered region. Initially, the center of the grasp pose is selected as the start point. However, the center point may not be suitable for the push action as the center point will always lie on the top of some object location. To find a suitable starting point for the push action, we traverse backward along the push vector with some small step values. As soon as we find a point having its depth value significantly higher (in camera frame) than the depth value of the center point, it is assigned as the start point for push action. It will allow the gripper to enter at a suitable depth level such that it can easily push the clutter along the push vector. The freest point is computed by applying a distance transform operation on the binary image of the input scene (Fig. \ref{fig:ci_final_declutter}). At any point in the image, the distance transform is defined as the minimum distance between the point and scene objects along with the inclusion of the robot workspace boundaries. The values in the distance transform are close to zero for all points near the objects and robot workspace boundaries and increase monotonically as we move away from both objects occupied region and robot workspace boundaries. The freest points would be the one having the highest value in the distance transform map (Fig. \ref{fig:ci_final_declutter}- \ref{fig:ci_binary}). 
	The objects are then pushed by the robot from the clutter zone in the direction towards the calculated freest point (Fig. \ref{fig:free_point_space}), effectively driving the objects to a new state to lessen the clutter state and increasing the overall grasp access of the objects.

\subsection{Improved and Adaptive Grasp Planning} \label{sec-improved-grasp-planning}

	\subsubsection{Baseline Grasp planning method}\label{sec:gdi}
	The domain-independent grasping method \cite{pharswan2019domain} is suitable for grasp planning of previously unseen novel objects. 
	First, it samples the grasp poses in the entire visible workspace and filters them based on the depth difference relative to the background (known a priori). Then, the remaining poses are grouped using k-means clustering and a final grasp pose is predicted for each cluster.  
	A Grasp Decide Index (GDI) is used to rank the resultant grasp poses (depicted in Fig. \ref{fig:GDI1}). More the value of the GDI index, the better is the grasp pose. The generated grasp pose configurations of the gripper are shown in  Fig. \ref{fig:overview}  as rectangles at the computed angle with a fixed-length equivalent to gripper opening. For more details, readers are advised to refer to the corresponding paper \cite{pharswan2019domain}.

	There are three major challenges to this method that have been found by experimentation. The first is clutter, due to which either the gripper collides with nearby objects resulting in a grasp failure, or multiple objects are picked together in a single grasp attempt. The latter case is not desirable if the objects need to be sorted post grasping. We address the issue of clutter by adding a clutter removal method in the grasping pipeline which is already described in the Sec. \ref{sec-clutter-description}.  
	Secondly, GDI index calculation is crucial to the baseline algorithm and there are many hyper-parameters that were set heuristically. We provide an empirical basis for setting these hyper-parameters appropriately, discussed further in the Sec. \ref{sec-gdi-optim}. 
	Lastly, the number of clusters for k-means was set to a fixed number. This can be a problem when the number of objects present in the scene is either very less or very high compared to the fixed value of k. We present a simple method to roughly estimate the number of objects present in the scene and equate it to the number of clusters in the k-means (Sec. \ref{sec-k-estimation}). The k-estimation method does not cause any significant time overhead while giving the performance boost of around $3\%$.  

\subsubsection{GDI optimization} \label{sec-gdi-optim}
\begin{figure}[!b]
	\centering
	\begin{subfigure}{0.22\textwidth}
		\centering
		\includegraphics[scale=0.30]{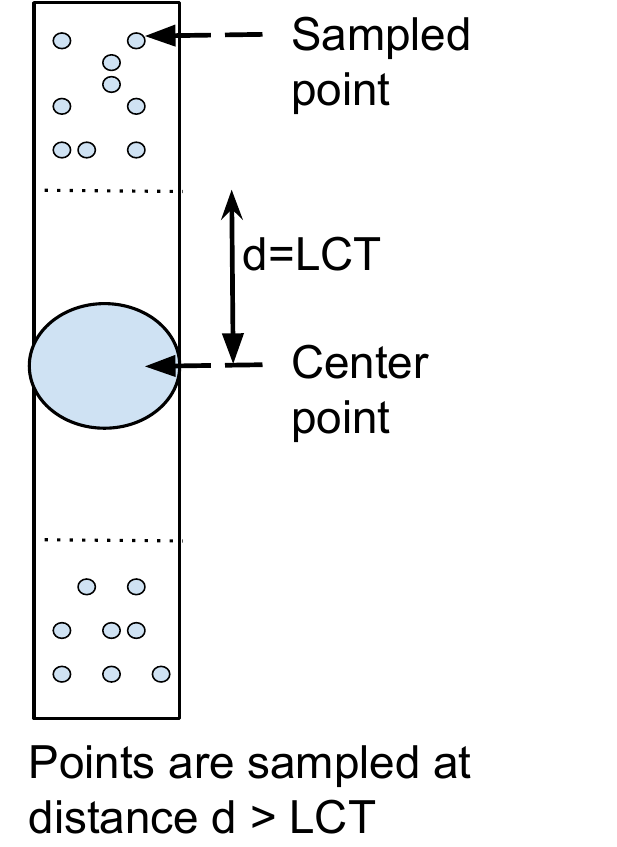}
		\caption{Points sampling}
		\label{fig:GDI1a}
	\end{subfigure}
	\begin{subfigure}{0.22\textwidth}
		\centering
		\includegraphics[scale=0.30]{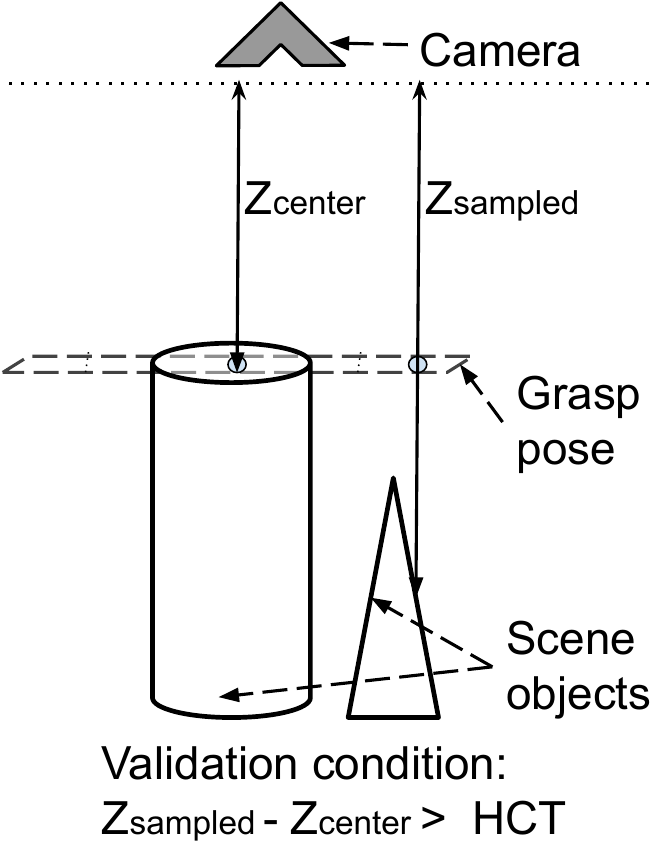}
		\caption{Points validation}
		\label{fig:GDI1b}
	\end{subfigure}
	\caption{Depiction of GDI Index calculation. (a) The  points  are  selected  only  if  the sampled  points  are  at  least  $\texttt{LCT}$  distance  away  from  the center point of the pose. (b) Points are valid when corresponding  depth  value  of  sampled  points is greater by at-least $\texttt{HCT}$ units from the depth value of the center point}
	\label{fig:GDI1}
\end{figure}

    GDI index is used to rank the candidate grasp poses.   
For calculating it, two threshold parameters are named herein as lateral-clearance-threshold ($\texttt{LCT}$) and height-clearance-threshold ($\texttt{HCT}$), which are very critical as it directly affects the GDI scores, which in turn affects the overall grasping performance. In Fig. \ref{fig:GDI1}, the role of these thresholds in the GDI calculation is depicted.
	The first threshold $\texttt{LCT}$ decides the minimum distance at which points can be sampled in the grasp pose rectangle (Fig. \ref{fig:GDI1a}).
	The second threshold $\texttt{HCT}$ decides whether a sampled point will contribute to the GDI score or not. If the corresponding depth value (in camera frame) of a sampled point (\texttt{Z$_{sampled}$}) is greater by at least $\texttt{HCT}$ units from the depth value of the center point (\texttt{Z$_{center}$}) then the sampled point is considered valid and will contribute to the GDI score  (Fig. \ref{fig:GDI1b}). Otherwise, it indicates that the gripper might collide with some object at this point while attempting the grasp action. Intuitively, the value of $\texttt{LCT}$ should be at least greater than the width of the target object at the grasping location, otherwise, it will falsely report the collision with the target object itself. The value of $\texttt{HCT}$ should be roughly equal to the depth difference till which the gripper would enter beneath the object for grasping it or in other words to the distance between the gripper endpoint and the target object surface at the grasp location at the time when gripper would close its finger to attempt the grasp. In Sec. \ref{sec-gdi-optim-results}, results of the empirical study are discussed which support the above-mentioned intuitions.

\subsubsection{Estimation of $k$ in k-means} \label{sec-k-estimation}
Algorithmically, the value of $\texttt{k}$ in k-means should be roughly equal to the number of objects present in the scene. It is not necessary that each object would be part of a separate cluster if $\texttt{k}$ is set exactly to the number of objects. Thus, the exact estimation of the number of objects is not crucial and a simple method is enough that only gives a rough estimate of $\texttt{k}$. One other motivation for choosing a simple method for the k-estimation is the time complexity of the method. An advanced method might be used for estimating the number of objects present in the scene, however, it will not add any significant improvements. In Sec. \ref{sec-analysis-k-means}, we present an ablation study to support our arguments. Our k-estimation method makes use of the two quantities, namely estimation of the area of spread of the objects in the workspace and the global clutter index. 
Area spread is estimated as the fraction of depth filtered pixels (i.e. belongings to the scene objects) which is already obtained at an intermediate step in the grasp planning algorithm. Global clutter index is obtained from our clutter estimation step described in Sec. \ref{sec-clutter-description}.  
A linear regression model is trained to estimate the number of objects ($k$) given the estimated area ($A$) and global clutter score ($G_{cs}$) of the spread: 
\begin{equation}\label{eq-k-means}
\mathit{k}=\beta_0 + \beta_1A +\beta_2G_{cs}
\end{equation}

For the given set of objects, the area spread might differ depending upon whether the objects are spread or tightly packed. The global clutter index becomes the deciding factor in such situations. 
As discussed before, a rough estimate is enough for the value of $\texttt{k}$. The weights parameters of Eq. \ref{eq-k-means} need to recalculate only when there is a dramatic change in the sizes of the objects present in the scene.

\begin{table*}[!t]
	\begin{center}
		\caption{Comparing different model variants based on the success rate, multi-picks-count  and mean-pick-per-hour. Evaluation metrics are defined in Sec. \ref{sec-eval-matrices}. The definitions of model names and other details are given in Sec. \ref{sec:compare}.  
        }
		
		\label{table:compare}
		\begin{tabular}{l|c c c| |c c c c} 
			  & \multicolumn{3}{c||}{\textbf{Simulation}} & \multicolumn{4}{c}{\textbf{Real World}} \\
			\hline
			Model Variant  & \makecell{GS \\ Grasp Success\\ Rate} & \makecell{MPC\\ (Multi Picks\\ Count(\%))} & \makecell{GS$_{wm}$ \\ Grasp success\\ rate} & \makecell{GS \\ Grasp Success\\ Rate} & \makecell{MPC\\ Multi Picks \\Count(\%)}  &  \makecell{GS$_{wm}$ \\ Grasp success\\rate}  & \makecell{MPPH \\ Mean Picks \\Per Hour} \\ 
			\hline
			\hline
			Baseline \cite{pharswan2019domain}   & 75.2  & 3.7 & 71.5  & 79.0  &  9.0 & 70.0 & 131 \\
			DexNet-2.0 \cite{dexnet2}  & 71.9 & 3.3 & 68.6 & 74.0  & 7.0 & 67.0 & 134 \\
			DexNet-4.0-PJ  \cite{dexnet4} & 76.2 & 2.6 & 73.6 &  80.0  & 6.0 & 74.0 & 144 \\
			
			Grasp-optim (Ours) & 77.5  & 3.5  & 74.0 & 81.0 & 8.0 & 73.0 & 142 \\
			
			Grasp-optim-adaptive (Ours)  & 80.9 & 1.9 & 79.0 & 84.0  & 6.0 & 78.0 & 150 \\
			

			\textbf{Disperse+Grasp (Ours)}   & \textbf{84.2} & \textbf{0.4} & \textbf{83.8} & \textbf{90.0}      & \textbf{1.0} & \textbf{89.0} & \textbf{159}\\
			\hline
			
		\end{tabular}
	\end{center}
\end{table*}

	\section{Results and Discussions}\label{sec:result}
	\subsection{Experimental setup}\label{sec_env_setup}
\begin{figure}[!t]
	\centering
	\begin{subfigure}{0.21\textwidth}
		\centering
		\includegraphics[scale=0.178]{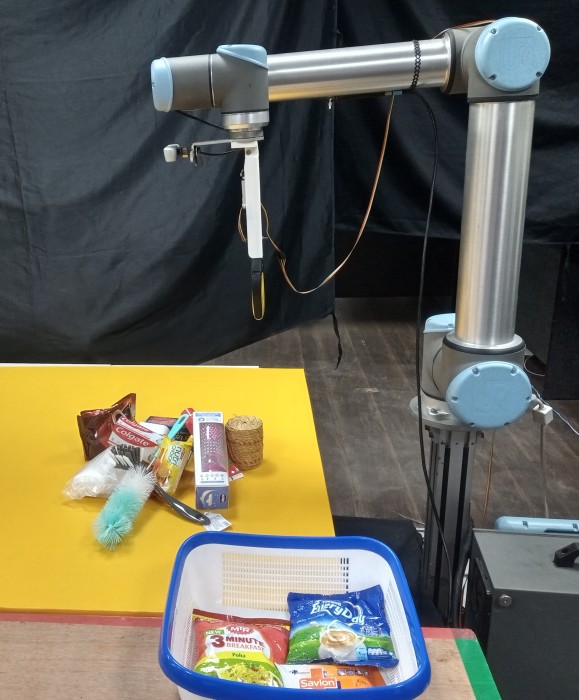}
		\caption{Real-world}
		\label{fig:env_real}
	\end{subfigure}
	\begin{subfigure}{0.21\textwidth}
		\centering
		\includegraphics[scale=0.2]{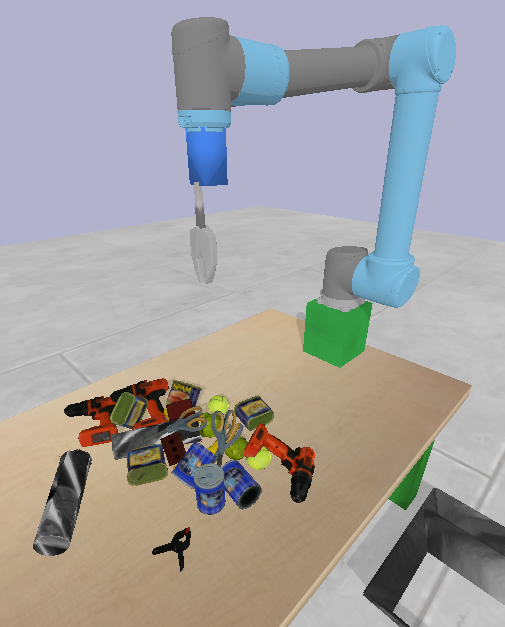}
		\caption{Simulation}
		\label{fig:env_sim}
	\end{subfigure}
	\caption{Environment setup consists of a robot arm with a gripper, an eye-in-hand camera and cluttered objects on the table.}
	\label{fig:env}
\end{figure}

	An experimental test-bed has been created to run experiments with a UR10 robotic arm,  consumer goods of various shapes, sizes \& materials, a RealSense RGBD camera, and a hand-designed two-finger gripper (Fig. \ref{fig:env_real}). To run multiple batch experiments for assessing the effect of various parameters over the grasping performance, a simulation environment is created that resembles the real-world setup using PyBullet (Fig. \ref{fig:env_sim}). In the simulation, a subset of YCB objects \cite{calli2015ycb} was used that consists of daily-use consumer goods. Results discussed in this section are obtained over a total of 2000 grasp trials in the simulation and 100 grasp trials in the Real-world for each presented method variant. For the experiments, a clutter of a maximum of 20 objects is created in the robot-workspace. The robot grasps the objects one by one till the workspace is empty. Then a novel random clutter is created again and the procedure is repeated till the total number of grasp attempts reaches the maximum number of trials to be performed. For each grasp trial, we record the following variables - the number of objects present in the scene before the grasp attempt, a binary variable indicating grasp success or failure, a binary variable indicating whether multiple objects are grasped simultaneously, global clutter score of the scene and the local clutter score concerning the target grasp pose.

	\subsection{Evaluation metrics} \label{sec-eval-matrices}
	The definitions of evaluation metrics used for the different experiments are given below. Shorthands are given in the brackets.\\
	Grasp Success (GS): Percentage of successful grasp over the total number of grasp attempts. Higher is better. \\
	Multi-Picks-Count (MPC): Total number of times when multiple objects are picked together while attempting for a grasp. Lower is better. \\
	Grasp Success (GS$_{wm}$): Grasp success rate when multi picks are counted as a failure. Higher is better. \\
    Mean-pick-per-hour (MPPH): Average number of successful pick and place in an hour time. Higher is better.

\subsection{Comparative Analysis of the proposed framework}\label{sec:compare}
In this section, a comparative analysis is given that evaluates the relative efficacy of different variants of our proposed framework along with the baseline method \cite{pharswan2019domain}, deep learning-based state-of-the-art grasping frameworks DexNet-2.0 \cite{dexnet2} and DexNet-4.0-PJ  \cite{dexnet4}.
The comparison of our method with the DexNet frameworks is done in the sense of universal picking. It means that the method should work well for the novel type of previously unseen objects and should be minimally sensitive to the camera, gripper, etc. For that purpose, the open-sourced learned models of DexNet frameworks are directly used without any retraining. Our experimental setup is similar to that used in the original work of DexNet (i.e. tabletop grasping in a heap of daily-use goods objects). The distance between the depth camera and the table surface is kept between 50 to 70 cm similar to the setup of the DexNet training environment. In the case of DexNet-4.0 \cite{dexnet4}, an ambidextrous gripping mechanism was used in their paper and two separate models are learned, one for a two-fingered parallel-jaw gripper and one for a suction gripper. For fair comparisons, the model trained with the parallel-jaw gripper (hence named as DexNet-4.0-PJ) is used herein. 

The short-hands for different variants and their specifications are as follows:\\
Baseline: Re-implementation of the baseline method. All the settings are kept as used in the method \cite{pharswan2019domain} \\
Grasp-optim: The grasp decide index (i.e. GDI) is optimized as described in Sec. \ref{sec-gdi-optim}. This setting is also used for all the subsequent variants.\\

Grasp-optim-adaptive: Number of clusters for k-means algorithms used in our grasp planning are not fixed but estimated using our k-estimation method (Also Referred as Grasp-only in the Sec. \ref{sec-ablation}).\\
Disperse+Grasp: All the settings are the same as the previous variant, additionally our proposed clutter estimation and removal method is used.

The comparative results are given in the Tab. \ref{table:compare}. 
Compared to the baseline, our grasp planning method has achieved around 5-6$\%$ gain in success-rate (GS) and 8-9$\%$ gain in success-rate (GS$_{wm}$) when multi-picks are counted as a failure. The setting of hyperparameters intelligently and the estimation of the number of clusters for k-means enabled the algorithm to provide a much better grasp pose. The MPC index is still significant for our grasp planning which is not desirable. 
Our final best framework, namely Disperse+Grasp, outperforms all the compared methods with a significant margin. It has achieved around 9-10$\%$ gain in success-rate (GS) and around $10\%$ relative gain in MPPH index compare to the deep-learning-based framework DexNet-4.0-PJ. A large part of the improved performance comes from the clutter removal part of the method. The MPC index is significantly reduced that increases the reliability and robustness of the method. The presented method has advantages both in terms of deployment efforts (data collection, training, etc.) and system requirements (computing power, memory, etc.) compared to the DexNet framework.

\subsection{Analysis of improvements in grasp planning algorithm }	

 \begin{figure}[!b]
	\centering
	\includegraphics[scale=0.5]{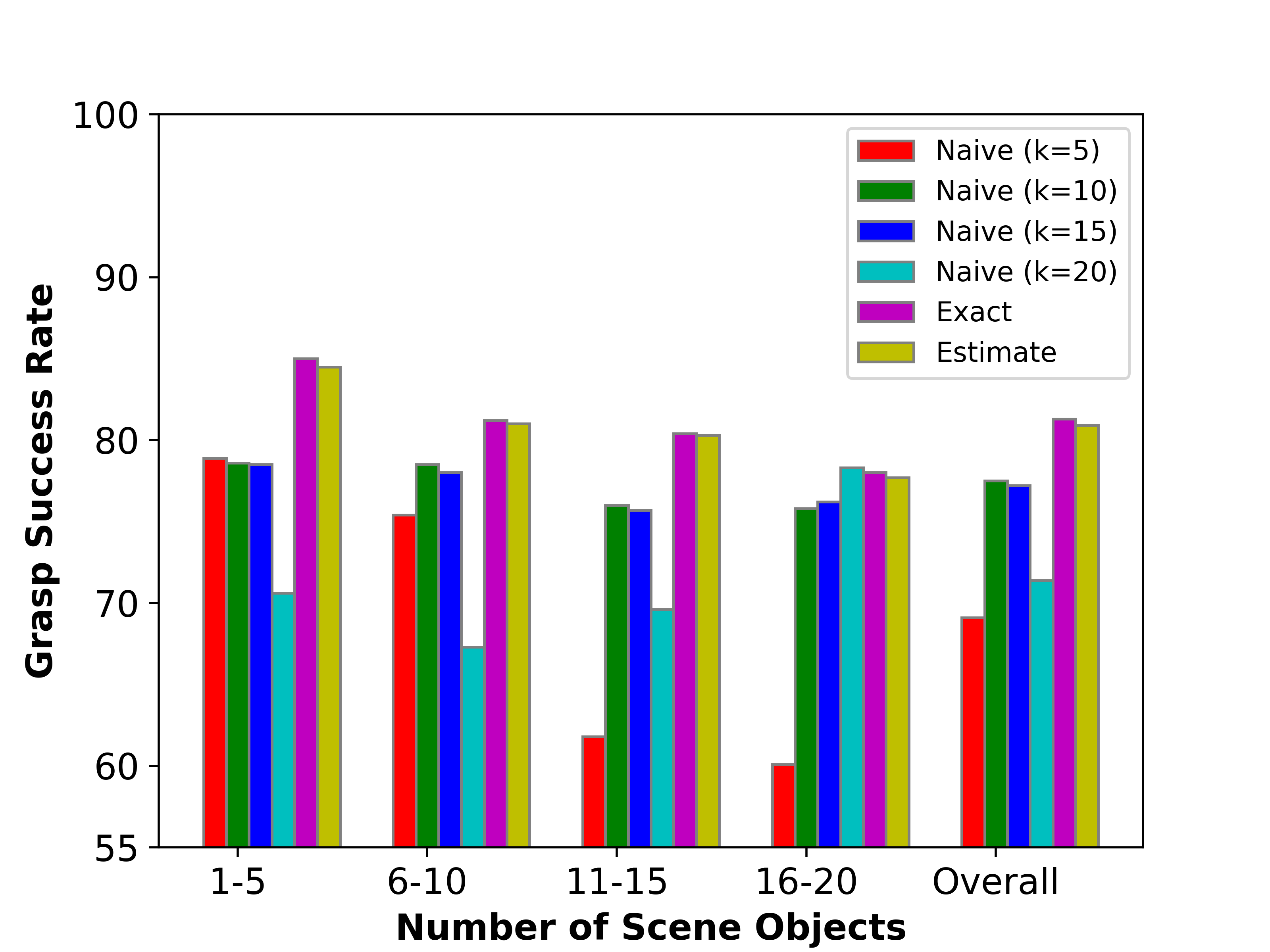}
	\caption{Analysis of the k-estimation method: Grasp Success Rate (GS) plotted for different variants. For details please refer to Sec. \ref{sec-analysis-k-means}}
	\label{fig:analysis-k-means}
\end{figure}
\subsubsection{K-estimation} \label{sec-analysis-k-means}
    For estimation of $k$ in the k-means algorithm used in the grasp planning method, a simple linear regression method was presented in Sec. \ref{sec-k-estimation} two things, first, a single fix value of $k$ is not optimal for varying number of objects in the scene. Second, our k-estimation method that roughly estimates the number of objects present in the scene is enough to provide the optimal performance. The results of this study are depicted in Fig. \ref{fig:analysis-k-means}. The y-axis of the figure represents the grasp success rate. Three different variants of the grasp planning algorithm are taken into consideration. First variants is named \texttt{Naive},  uses the fix number of clusters (i.e. k = 5, 10, 15 and 20). The second variant \texttt{Exact}, equates the number of clusters with the number of objects currently present in the scene in each run of the algorithm. The ground truth information of the number of objects is directly accessed from the known simulation state. The third variant \texttt{Estimate} is our k-estimation method.  
    
    From the results depicted in Fig. \ref{fig:analysis-k-means}, it can be understood that a rough estimate of the number of objects is enough to get the performance compared with any advanced method for estimating the number of objects in the scene. For our method, the grasping success rate is high for all the cases with the varying number of objects in the scene. The performance improves a little when the exact number of objects are known at each run of the algorithm, however, this is not possible in a real-world setting. The naive approach of setting a fixed number for the value of $k$ performs well only when the number of objects present in the scene is in a nearby range of $k$. 
    
   	\begin{figure}[!b]
	\centering
	\includegraphics[scale=0.5]{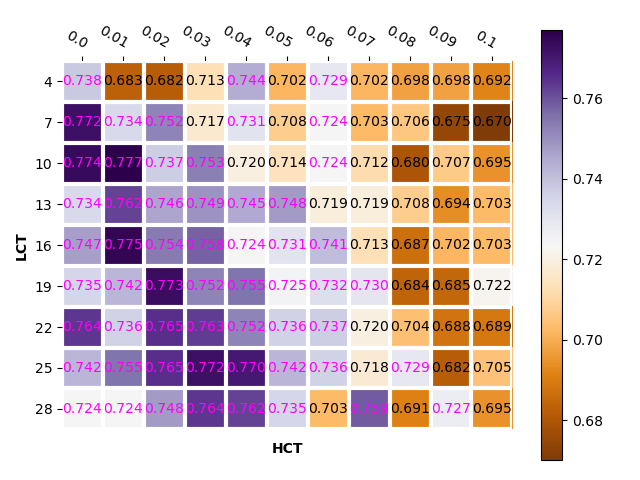}
	\caption{GDI optimization: Heat map of grasp success rate for different values of thresholds $\texttt{LCT}$ and $\texttt{HCT}$. For details please refer to Sec.  \ref{sec-gdi-optim-results}.}
	\label{fig:heat_map}
\end{figure}
    
\subsubsection{GDI optimization} \label{sec-gdi-optim-results}
	An empirical study is done to study the effects of the crucial threshold hyperparameters used in GDI index calculation. Fig. \ref{fig:heat_map} shows the heat-map of grasp success rate generated from the results of the batch experiments for different values of the hyperparameters described in Sec. \ref{sec-gdi-optim}.
	It can be noticed from the results in Fig. \ref{fig:heat_map} that there is a particular pattern in the heat-map except a few spurious entries. The blue color represents the higher success rate and the brown color represents the opposite. It can be deducted that for threshold $\texttt{LCT}$, the optimal value lies somewhere between 7 to 25 (in pixel units), and for threshold $\texttt{HCT}$, the optimal value lies between 0.0 to 0.04 (in meters). The average width of the target objects calculated at grasp locations is found to be 16 in pixel units. 
	If the value of $\texttt{LCT}$ is too low ($<<$ average width of the objects), then most likely, the GDI will detect the gripper collision with the object of interest itself which can be termed as false-positive. If the value of $\texttt{LCT}$ is too high then the GDI collision detection window is very small. In that case, having a higher value for $\texttt{HCT}$ keeps the collision detection strategy stricter and thus avoids false negatives. The pattern in the Fig. \ref{fig:heat_map} confirms the same.

\subsection{An analysis over the effect of clutter}
\label{sec-ablation}
\begin{figure}[!th]
		\centering

		\begin{subfigure}{0.155\textwidth}
			\captionsetup{justification=centering}

			\centering
			\includegraphics[width=\textwidth]{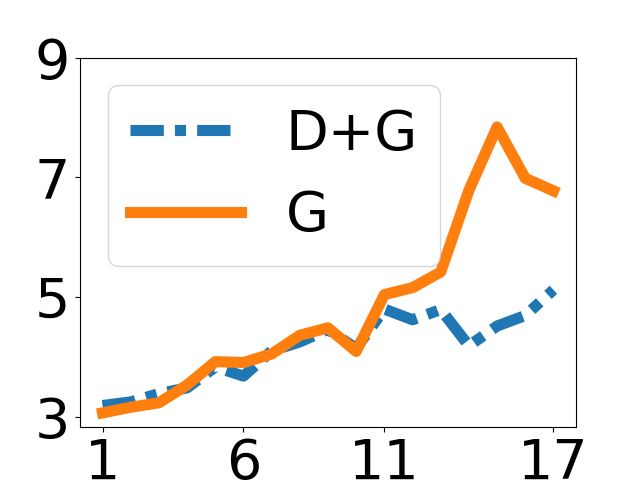}
			\caption{Local clutter score}
			 \label{fig:clutter-compare}
		\end{subfigure}
		\begin{subfigure}{0.155\textwidth}
				\captionsetup{justification=centering}

			\centering
			\includegraphics[width=\textwidth]{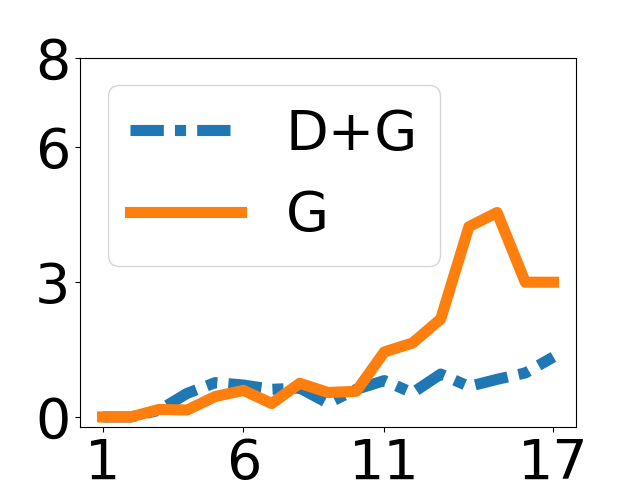}
			\caption{MPC}
			\label{fig:double-compare}
		\end{subfigure}
		\begin{subfigure}{0.155\textwidth}
			\captionsetup{justification=centering}

			\centering
			\includegraphics[width=\textwidth]{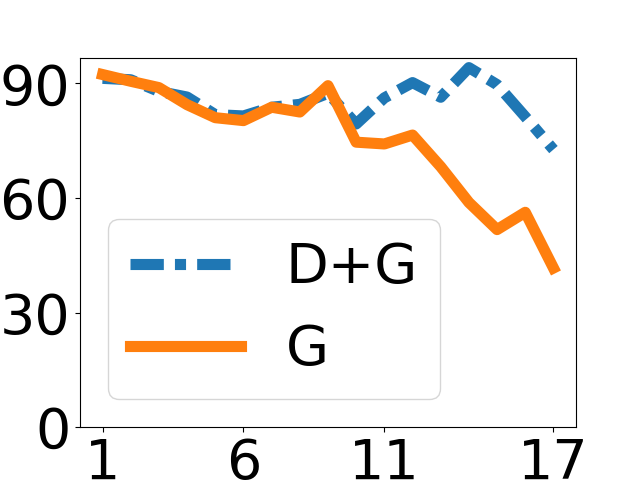}
			\caption{Success rate
			}
			\label{fig:suc-compare}
		\end{subfigure}
		
		\caption{An analysis to asses the effect of clutter over the performance of grasping algorithm and how the clutter removal framework helps reduce the problem. Here \texttt{D+G} represents \texttt{Disperse+Grasp} method and \texttt{G} represents \texttt{Grasping-only} method. X-axis represents the number of objects present in the scene during a grasp attempt. For details please refer to Sec. \ref{sec-ablation}.}
		\label{fig:multi-picked-analysis}
	\end{figure}

To assess empirically, how clutter causes the increment in grasp failures and in the MPC index, an analysis is done. In this, ours Disperse+Grasp method and Grasp-only method are compared (Fig. \ref{fig:multi-picked-analysis}). Experiment results consisting of 2000 grasp trials were segregated into different bins with respect to the number of objects present in the scene during a grasp attempt. First, the average local clutter score is calculated in each bin. It is obtained by taking the average local clutter score with respect to the target grasp pose across all the experiments in the particular bin. The local clutter score corresponding to a grasp pose is measured before the grasp attempt. As expected for the Grasp-only method, the clutter score increases with the increase in the number of objects present in the scene at the time of grasp action (Fig. \ref{fig:clutter-compare}). Whereas, for the Disperse+Grasp method clutter score increases relatively slower. Because in this method the algorithm chooses push-to-move action instead of a grasp action when the clutter score is above a threshold. In Fig. \ref{fig:double-compare}, percentage of MPC (with respect to the total number of grasp trials in the particular bin) and in Fig. \ref{fig:suc-compare}, Grasp Success Rate (GS) are plotted. For the Grasp-only method, with the increase of clutter (i.e. more number of objects in the scene), the MPC increases and the Grasp Success decreases, which is expected. Whereas, the Disperse+Grasp method is able to maintain the performance level despite the increase in clutter. 
	
	\section{Conclusions}
	In this paper, we presented a framework to disperse the objects for better singulation so that the objects can be picked by the robot with certainty. With the use of this framework, the cases of multi-picks (i.e MPC index) and the grasp failures due to lack of space for the gripper to extend are reduced. For grasp planning, we have proposed an improved version of a recent domain-independent grasping method. The proposed disperse framework can be used with any grasp planning framework. Failure cases of our framework include objects going out of the robot workspace, objects slipping from the gripper after being grasped due to complex geometry, and predicted grasp pose that happened to be at the corner side of the object.   
	For future works, more types of action primitives can be explored which can more efficiently singulate the objects. Also, the framework can be extended to include grasping strategies for the deformable and fragile objects with the use of some extra sensors such as tactile-sensor.
	
	\bibliographystyle{./bibliography/IEEEtran}
	\bibliography{ref.bib}

\begin{thebibliography}{10}
\providecommand{\url}[1]{#1}
\csname url@samestyle\endcsname
\providecommand{\newblock}{\relax}
\providecommand{\bibinfo}[2]{#2}
\providecommand{\BIBentrySTDinterwordspacing}{\spaceskip=0pt\relax}
\providecommand{\BIBentryALTinterwordstretchfactor}{4}
\providecommand{\BIBentryALTinterwordspacing}{\spaceskip=\fontdimen2\font plus
\BIBentryALTinterwordstretchfactor\fontdimen3\font minus
  \fontdimen4\font\relax}
\providecommand{\BIBforeignlanguage}[2]{{%
\expandafter\ifx\csname l@#1\endcsname\relax
\typeout{** WARNING: IEEEtran.bst: No hyphenation pattern has been}%
\typeout{** loaded for the language `#1'. Using the pattern for}%
\typeout{** the default language instead.}%
\else
\language=\csname l@#1\endcsname
\fi
#2}}
\providecommand{\BIBdecl}{\relax}
\BIBdecl

\bibitem{stoica2017berkeley}
I.~Stoica, D.~Song, R.~A. Popa, D.~Patterson, M.~W. Mahoney, R.~Katz, A.~D.
  Joseph, M.~Jordan, J.~M. Hellerstein, J.~E. Gonzalez \emph{et~al.}, ``A
  berkeley view of systems challenges for ai,'' \emph{arXiv preprint
  arXiv:1712.05855}, 2017.

\bibitem{amazon_picking_challenge}
N.~Correll, K.~E. Bekris, D.~Berenson, O.~Brock, A.~Causo, K.~Hauser, K.~Okada,
  A.~Rodriguez, J.~M. Romano, and P.~R. Wurman, ``Analysis and observations
  from the first amazon picking challenge,'' \emph{IEEE Transactions on
  Automation Science and Engineering}, vol.~15, no.~1, pp. 172--188, 2018.

\bibitem{reinhart2011automatic}
G.~Reinhart, S.~H{\"u}ttner, and S.~Krug, ``Automatic configuration of robot
  systems-upward and downward integration,'' in \emph{International Conference
  on Intelligent Robotics and Applications}.\hskip 1em plus 0.5em minus
  0.4em\relax Springer, 2011, pp. 102--111.

\bibitem{el2019simulation}
M.~El-Shamouty, K.~Kleeberger, A.~L{\"a}mmle, and M.~Huber, ``Simulation-driven
  machine learning for robotics and automation,'' \emph{tm-Technisches Messen},
  vol.~86, no.~11, pp. 673--684, 2019.

\bibitem{pharswan2019domain}
S.~V. Pharswan, M.~Vohra, A.~Kumar, and L.~Behera, ``Domain-independent
  unsupervised detection of grasp regions to grasp novel objects,'' in
  \emph{2019 IEEE/RSJ International Conference on Intelligent Robots and
  Systems (IROS)}.\hskip 1em plus 0.5em minus 0.4em\relax IEEE, 2019, pp.
  640--645.

\bibitem{kleeberger2020survey}
K.~Kleeberger, R.~Bormann, W.~Kraus, and M.~F. Huber, ``A survey on
  learning-based robotic grasping,'' \emph{Current Robotics Reports}, pp.
  1--11, 2020.

\bibitem{jo2020object}
H.~Jo and J.-B. Song, ``Object-independent grasping in heavy clutter,''
  \emph{Applied Sciences}, vol.~10, no.~3, p. 804, 2020.

\bibitem{vohra2019real}
M.~Vohra, R.~Prakash, and L.~Behera, ``Real-time grasp pose estimation for
  novel objects in densely cluttered environment,'' in \emph{2019 28th IEEE
  International Conference on Robot and Human Interactive Communication
  (RO-MAN)}.\hskip 1em plus 0.5em minus 0.4em\relax IEEE, 2019, pp. 1--6.

\bibitem{dexnet2}
J.~{Mahler}, J.~{Liang}, S.~{Niyaz}, M.~{Laskey}, R.~{Doan}, X.~{Liu},
  J.~{Aparicio}, and K.~{Goldberg}, ``Dex-net 2.0: Deep learning to plan robust
  grasps with synthetic point clouds and analytic grasp metrics,'' in
  \emph{Robotics: Science and Systems 2017}, vol.~13, 2017.

\bibitem{dexnet4}
J.~Mahler, M.~Matl, V.~Satish, M.~Danielczuk, B.~DeRose, S.~McKinley, and
  K.~Goldberg, ``Learning ambidextrous robot grasping policies,'' \emph{Science
  Robotics}, vol.~4, no.~26, 2019.

\bibitem{tremblay2018deep}
J.~Tremblay, T.~To, B.~Sundaralingam, Y.~Xiang, D.~Fox, and S.~Birchfield,
  ``Deep object pose estimation for semantic robotic grasping of household
  objects,'' in \emph{Conference on Robot Learning}.\hskip 1em plus 0.5em minus
  0.4em\relax PMLR, 2018, pp. 306--316.

\bibitem{levine2018learning}
S.~Levine, P.~Pastor, A.~Krizhevsky, J.~Ibarz, and D.~Quillen, ``Learning
  hand-eye coordination for robotic grasping with deep learning and large-scale
  data collection,'' \emph{The International Journal of Robotics Research},
  vol.~37, no. 4-5, pp. 421--436, 2018.

\bibitem{kalashnikov2018qt}
D.~Kalashnikov, A.~Irpan, P.~Pastor, J.~Ibarz, A.~Herzog, E.~Jang, D.~Quillen,
  E.~Holly, M.~Kalakrishnan, V.~Vanhoucke \emph{et~al.}, ``Qt-opt: Scalable
  deep reinforcement learning for vision-based robotic manipulation,'' in
  \emph{Conference on Robot Learning}.\hskip 1em plus 0.5em minus 0.4em\relax
  PMLR, 2018.

\bibitem{james2019sim}
S.~James, P.~Wohlhart, M.~Kalakrishnan, D.~Kalashnikov, A.~Irpan, J.~Ibarz,
  S.~Levine, R.~Hadsell, and K.~Bousmalis, ``Sim-to-real via sim-to-sim:
  Data-efficient robotic grasping via randomized-to-canonical adaptation
  networks,'' in \emph{Proceedings of the IEEE/CVF Conference on Computer
  Vision and Pattern Recognition}, 2019, pp. 12\,627--12\,637.

\bibitem{redmon2015real}
J.~Redmon and A.~Angelova, ``Real-time grasp detection using convolutional
  neural networks,'' in \emph{2015 IEEE International Conference on Robotics
  and Automation (ICRA)}.\hskip 1em plus 0.5em minus 0.4em\relax IEEE, 2015,
  pp. 1316--1322.

\bibitem{kumar2019semi}
A.~Kumar and L.~Behera, ``Semi supervised deep quick instance detection and
  segmentation,'' in \emph{2019 International Conference on Robotics and
  Automation (ICRA)}.\hskip 1em plus 0.5em minus 0.4em\relax IEEE, 2019, pp.
  8325--8331.

\bibitem{zeng2020tossingbot}
A.~Zeng, S.~Song, J.~Lee, A.~Rodriguez, and T.~Funkhouser, ``Tossingbot:
  Learning to throw arbitrary objects with residual physics,'' \emph{IEEE
  Transactions on Robotics}, vol.~36, no.~4, pp. 1307--1319, 2020.

\bibitem{song2020grasping}
S.~Song, A.~Zeng, J.~Lee, and T.~Funkhouser, ``Grasping in the wild: Learning
  6dof closed-loop grasping from low-cost demonstrations,'' \emph{IEEE Robotics
  and Automation Letters}, vol.~5, no.~3, pp. 4978--4985, 2020.

\bibitem{zeng2018learning}
A.~Zeng, S.~Song, S.~Welker, J.~Lee, A.~Rodriguez, and T.~Funkhouser,
  ``Learning synergies between pushing and grasping with self-supervised deep
  reinforcement learning,'' in \emph{2018 IEEE/RSJ International Conference on
  Intelligent Robots and Systems (IROS)}.\hskip 1em plus 0.5em minus
  0.4em\relax IEEE, 2018, pp. 4238--4245.

\bibitem{patten_action_2018}
T.~Patten, M.~Zillich, and M.~Vincze, ``Action {Selection} for {Interactive}
  {Object} {Segmentation} in {Clutter},'' in \emph{2018 {IEEE}/{RSJ}
  {International} {Conference} on {Intelligent} {Robots} and {Systems}
  ({IROS})}, Oct. 2018, pp. 6297--6304, iSSN: 2153-0866.

\bibitem{katz_clearing_2013}
D.~Katz, M.~Kazemi, J.~A. Bagnell, and A.~Stentz, ``Clearing a pile of unknown
  objects using interactive perception,'' in \emph{2013 {IEEE} {International}
  {Conference} on {Robotics} and {Automation}}, May 2013, pp. 154--161, iSSN:
  1050-4729.

\bibitem{danielczuk_linear_2018}
M.~Danielczuk, J.~Mahler, C.~Correa, and K.~Goldberg, ``Linear {Push}
  {Policies} to {Increase} {Grasp} {Access} for {Robot} {Bin} {Picking},'' in
  \emph{2018 {IEEE} 14th {International} {Conference} on {Automation} {Science}
  and {Engineering} ({CASE})}, Aug. 2018, pp. 1249--1256, iSSN: 2161-8089.

\bibitem{cluster-push}
Z.~Dong, S.~Krishnan, S.~Dolasia, A.~Balakrishna, M.~Danielczuk, and
  K.~Goldberg, ``Automating planar object singulation by linear pushing with
  single-point and multi-point contacts,'' in \emph{2019 IEEE 15th
  International Conference on Automation Science and Engineering (CASE)}, 2019,
  pp. 1429--1436.

\bibitem{imran_singulation_2019}
\BIBentryALTinterwordspacing
A.~Imran, S.-H. Kim, Y.-B. Park, I.~H. Suh, and B.-J. Yi,
  ``\BIBforeignlanguage{en}{Singulation of {Objects} in {Cluttered}
  {Environment} {Using} {Dynamic} {Estimation} of {Physical} {Properties}},''
  \emph{\BIBforeignlanguage{en}{Applied Sciences}}, vol.~9, no.~17, p. 3536,
  Jan. 2019, number: 17 Publisher: Multidisciplinary Digital Publishing
  Institute. [Online]. Available:
  \url{https://www.mdpi.com/2076-3417/9/17/3536}
\BIBentrySTDinterwordspacing

\bibitem{khan_scattering_2020}
\BIBentryALTinterwordspacing
M.~U.~A. Khan, A.~Imran, S.~Kim, H.~Hwang, J.~Y. Lee, S.~Lee, and B.-J. Yi,
  ``\BIBforeignlanguage{en}{Scattering or {Pushing} for {Object} {Singulation}
  in {Cluttered} {Environment}: {Case} {Study} with {Soma} {Cube}},''
  \emph{\BIBforeignlanguage{en}{Applied Sciences}}, vol.~10, no.~24, p. 9153,
  Jan. 2020, number: 24 Publisher: Multidisciplinary Digital Publishing
  Institute. [Online]. Available:
  \url{https://www.mdpi.com/2076-3417/10/24/9153}
\BIBentrySTDinterwordspacing

\bibitem{gupta_using_2012}
M.~Gupta and G.~S. Sukhatme, ``Using manipulation primitives for brick sorting
  in clutter,'' in \emph{2012 {IEEE} {International} {Conference} on {Robotics}
  and {Automation}}, May 2012, pp. 3883--3889, iSSN: 1050-4729.

\bibitem{berg_crowding_2009}
\BIBentryALTinterwordspacing
R.~v.~d. Berg, F.~W. Cornelissen, and J.~B. T.~M. Roerdink,
  ``\BIBforeignlanguage{en}{A crowding model of visual clutter},''
  \emph{\BIBforeignlanguage{en}{Journal of Vision}}, vol.~9, no.~4, pp. 24--24,
  Apr. 2009, publisher: The Association for Research in Vision and
  Ophthalmology. [Online]. Available:
  \url{http://jov.arvojournals.org/article.aspx?articleid=2193442}
\BIBentrySTDinterwordspacing

\bibitem{yu_modeling_2014}
\BIBentryALTinterwordspacing
C.-P. Yu, D.~Samaras, and G.~J. Zelinsky, ``\BIBforeignlanguage{en}{Modeling
  visual clutter perception using proto-object segmentation},''
  \emph{\BIBforeignlanguage{en}{Journal of Vision}}, vol.~14, no.~7, pp. 4--4,
  Jun. 2014, publisher: The Association for Research in Vision and
  Ophthalmology. [Online]. Available:
  \url{http://jov.arvojournals.org/article.aspx?articleid=2194016}
\BIBentrySTDinterwordspacing

\bibitem{rosenholtz_feature_2006}
\BIBentryALTinterwordspacing
R.~Rosenholtz, Y.~Li, Z.~Jin, and J.~Mansfield,
  ``\BIBforeignlanguage{en}{Feature congestion: {A} measure of visual
  clutter},'' \emph{\BIBforeignlanguage{en}{Journal of Vision}}, vol.~6, no.~6,
  pp. 827--827, Jun. 2006, publisher: The Association for Research in Vision
  and Ophthalmology. [Online]. Available:
  \url{http://jov.arvojournals.org/article.aspx?articleid=2133829}
\BIBentrySTDinterwordspacing

\bibitem{calli2015ycb}
B.~Calli, A.~Singh, A.~Walsman, S.~Srinivasa, P.~Abbeel, and A.~M. Dollar,
  ``The ycb object and model set: Towards common benchmarks for manipulation
  research,'' in \emph{2015 international conference on advanced robotics
  (ICAR)}.\hskip 1em plus 0.5em minus 0.4em\relax IEEE, 2015, pp. 510--517.

\end{thebibliography}

\end{document}